\documentclass{article} 
\usepackage{iclr2023_conference,times}
\usepackage[linesnumbered,ruled,vlined]{algorithm2e}
\usepackage{hyperref}
\usepackage{url}
\usepackage[utf8]{inputenc} 
\usepackage[T1]{fontenc}    
\usepackage{hyperref}       
\usepackage{url}            
\usepackage{booktabs}       
\usepackage{amsfonts}       
\usepackage{nicefrac}       
\usepackage{microtype}      
\usepackage{xcolor}         
\usepackage{subfig,graphicx}
\usepackage{amsmath}
\usepackage{wrapfig}
\usepackage{amssymb}
\usepackage{multirow}
\usepackage{setspace}
\usepackage{caption}
\usepackage{xcolor}
\usepackage{multirow}
\usepackage{bbm}
\newcommand*{\method}{CAKE}
\usepackage{colortbl}
\definecolor{myyellow}{rgb}{1,1, 0.6}
\definecolor{myorange}{rgb}{1, 0.8, 0.6}
\definecolor{myred}{rgb}{1, 0.6, 0.6}

\title{CAusal and collaborative proxy-tasKs lEarning for Semi-Supervised Domain Adaptation}

\begin{document}

\maketitle

\begin{abstract}
Semi-supervised domain adaptation (SSDA) adapts a learner to a new domain by effectively utilizing source domain data and a few labeled target samples. It is a practical yet under-investigated research topic. In this paper, we analyze the SSDA problem from two perspectives that have previously been overlooked, and correspondingly decompose it into two \emph{key subproblems}: \emph{robust domain adaptation (DA) learning} and \emph{maximal cross-domain data utilization}.  \textbf{(i)}  From a causal theoretical view, a robust DA model should distinguish the invariant ``concept'' (key clue to image label) from the nuisance of confounding factors across domains. To achieve this goal, we propose to generate \emph{concept-invariant samples} to enable the model to classify the samples through causal intervention, yielding improved generalization guarantees; \textbf{(ii)} Based on the robust DA theory, we aim to exploit the maximal utilization of rich source domain data and a few labeled target samples to boost SSDA further. Consequently, we propose a collaboratively debiasing learning framework that utilizes two complementary semi-supervised learning (SSL) classifiers to mutually exchange their unbiased knowledge, which helps unleash the potential of source and target domain training data, thereby producing more convincing pseudo-labels. Such obtained labels facilitate cross-domain feature alignment and duly improve the invariant concept learning. In our experimental study, we show that
the proposed model significantly outperforms SOTA methods in terms of effectiveness and generalisability on SSDA datasets.
\end{abstract}

\section{Introduction}
\label{intro} Deep neural networks excel at learning from large labeled datasets in computer vision~\cite{li2020unsupervised,liu2021swin,han2022survey,zhang2019frame,li2022hero,zhang2021consensus,li2022dilated} and natural language processing tasks~\cite{feng2020language,jawahar2019does,kowsari2019text,lv2023ideal,li2022end,zhang2022magic} but struggle to generalize to new target domains. This limits their real-world utility, as it is impractical to collect a large new dataset for every new deployment domain~\cite{zhang2022boostmis}. To alleviate this problem, 
 Domain Adaptation (DA) is proposed that aims to transfer training knowledge to the new domain (\emph{target} $\mathcal{D}= \mathcal{D}_\mathcal{T}$) using the labeled data available from the original domain (\emph{source}  $\mathcal{D}= \mathcal{D}_\mathcal{S}$), which can alleviate the poor generalization of learned deep neural networks when the data distribution  significantly deviates from the original domain~\cite{wang2018deep,you2019universal,tzeng2017adversarial}. In the DA community,  recent works~\cite{saito2019semi}  
  have shown that the presence of few labeled data from the target domain can significantly boost the performance of  deep learning-based models. This observation led to the formulation of Semi-Supervised Domain Adaptation (SSDA), which is a variant of Unsupervised Domain Adaptation (UDA)~\cite{venkateswara2017deep} to facilitate model training with rich labels from $\mathcal{D}_\mathcal{S}$ and a few labeled samples from $\mathcal{D}_\mathcal{T}$. For the fact that we can easily collect such additional labels on the target data in real-world applications, SSDA has the potential to render the adaptation problem more practical and promising in comparison to UDA.

 
 Broadly, most contemporary approaches~\cite{ganin2016domain,jiang2020bidirectional,kim2020attract,yoon2022semi} handle the SSDA task based on two domain shift assumptions,  where $\mathcal{X}$ and $\mathcal{Y}$ respectively denote the samples and their corresponding labels: (i) \emph{Covariate Shift}, $P(\mathcal{X}|\mathcal{D}= \mathcal{D}_\mathcal{S}) \neq P(\mathcal{X}|\mathcal{D}= \mathcal{D}_\mathcal{T})$; (ii) \emph{Conditional Shift},  $P(\mathcal{Y}|\mathcal{X},\mathcal{D}= \mathcal{D}_\mathcal{S}) \neq P(\mathcal{Y}|\mathcal{X},\mathcal{D}= \mathcal{D}_\mathcal{T})$, refers to the difference of conditional label distributions of cross-domain data. Intuitively, one straightforward solution for SSDA is to learn the common features to mitigate the domain shift issues. Further quantitative analyses, however, indicate that the model trained with supervision on a few labeled target samples and labeled source data can just ensure partial cross-domain feature alignment~\cite{kim2020attract,zhang2020relational}. 
 That is, it only aligns the features of labeled target samples and their correlated nearby samples with the corresponding feature clusters in the source domain.


  \begin{figure*}[t]
\includegraphics[width=1\textwidth]{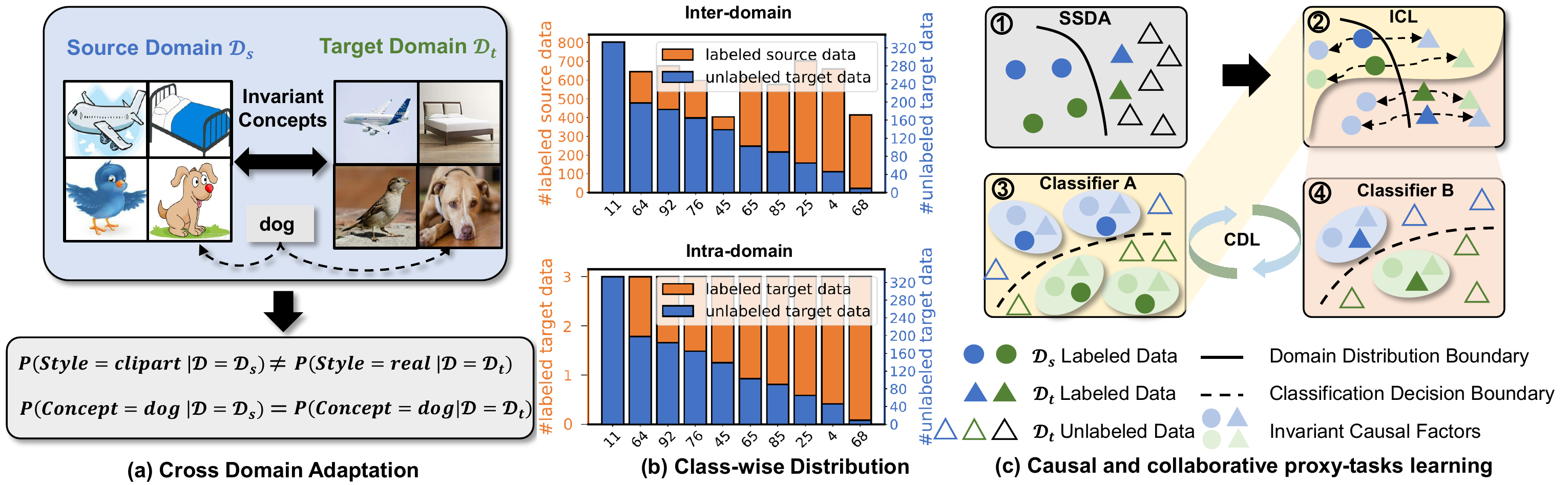}
\centering\caption{(a) Four DA cases (``Clipart'' $\rightarrow$ ``Real''). (b) Class-wise distribution of source domain and target domain. (c) A simplified version that indicates how our proposed model facilitates the SSDA.  }
\label{fig:intro}
\vspace{-0.3cm}
\end{figure*}

To systematically study the SSDA problem, we begin by asking two fundamental questions, \textbf{Q1}:\emph{What properties should a robust DA model have?} 
To answer this question, we first present a DA example in Figure~\ref{fig:intro}(a), which suggests that the image ``style'' in $\mathcal{D}=\mathcal{D}_\mathcal{T}$ is drastically different from the $\mathcal{D}=\mathcal{D}_\mathcal{S}$.
A classifier trained on the source domain may fail to predict correct labels even though the ``concept'' (\emph{e.g.}, plane) is invariant with a similar outline.
The truth is that the minimalist style features being invariant in ``clipart'' domain plays a critical factor in the trained classifier, which may consequently downplay the concept features simply because they are not as invariant as style features.
Importantly, such an observation reveals the fundamental reason of the two domain shift assumptions, \emph{i.e.}, $P(Style=clipart|\mathcal{D}=\mathcal{D}_\mathcal{S}) \neq P(Style=real|\mathcal{D}=\mathcal{D}_\mathcal{T})$. 
Therefore, a robust DA model needs to distinguish the invariant concept features in $\mathcal{X}$ across domains from the changing style. \textbf{Q2}: \emph{How to maximally exploit the target domain supervision for robust SSDA?} 
As discussed, supervised learning on the few target labels cannot guarantee the global cross-domain feature alignment, which hurts the model generalization for invariant learning. 
A commonly known approach in this few labeled setting, semi-supervised learning (SSL),
uses a trained model on labeled data to predict convincing pseudo-labels for the unlabeled data. 
This approach relies on the ideal assumption that the labeled and unlabeled data have the same marginal distribution of label over classes to generate pseudo-labels.
However, Figure~\ref{fig:intro}(b) indicates these distributions are different in both inter-domain and intra-domain. 
This may result in the imperfect label prediction that causes the well-known \emph{confirmation bias}~\cite{arazo2020pseudo},  affecting the model feature alignment capability. 
Further, in the SSDA setting, we have three sets of data, \emph{i.e.}, source domain data, labeled and unlabeled target domain data. 
One single model for SSDA may be hard to generalize to the three sets with different label distributions. 
Thus, the premise of better utilization of labeled target samples is to
mitigate undesirable bias and reasonably utilize the multiple sets.
Summing up, 
 these limitations call for reexamination of SSDA and its solutions.

To alleviate the aforementioned limitations,
we propose a framework called  \underline{\textbf{CA}}usal collaborative proxy-tas\underline{\textbf{K}}s l\underline{\textbf{E}}arning
(\textbf{\method{}}) which is illustrated in Figure~\ref{fig:intro}(c).
In the first step, we formalize the DA task using a causal graph. Then leveraging causal tools, we identify the "style" as the \emph{confounder} and derive the invariant concepts across domains. In the subsequent steps, we build two classifiers based on the invariant concept to utilize rich information from cross-domain data for better SSDA. In this way, \method{} explicitly decomposes the SSDA into two proxy subroutines, namely \emph{Invariant Concept Learning Proxy} (ICL) and \emph{Collaboratively Debiasing Learning Proxy} (CDL). In ICL, we identify the key to robust DA is that the underlying concepts are consistent across domains, and the \emph{confounder} is the style that prevents the model from learning the invariant concept (\emph{C}) for accurate DA. Therefore, a robust DA model should be an invariant predictor $P(\mathcal{Y}|\hat{\mathcal{X}}, \mathcal{D}= \mathcal{D}_\mathcal{T})=P(\mathcal{Y}|\hat{\mathcal{X}},\mathcal{D}= \mathcal{D}_\mathcal{S})$) under causal interventions. 
To address the problem, 
we devise a causal factor generator (CFG) that can produce concept-invariant  samples $\hat{\mathcal{X}}$ with different style to facilitate the DA model to effectively learn the invariant concept. 
As such, our ICL may be regarded as an improved version of Invariant Risk Minimization (IRM)~\cite{arjovsky2019invariant} for SSDA, which equips the
model with the ability to learn the concept features that are invariant to styles. 
In CDL, with the invariant concept learning
as the foundation, we aim to unleash the potential of three sets of cross-domain data for better SSDA. Specifically, we build two correlating and complementary pseudo-labeling based semi-supervised learning (SSL) classifiers for $\mathcal{D}_\mathcal{S}$ and $\mathcal{D}_\mathcal{T}$ with self-penalization. These two classifiers ensure that the mutual knowledge is exchanged to expand the number of ``labeled" samples in the target domain, thereby bridging the feature distribution gap. 
Further, to reduce the \emph{confirmation bias} learned from respective labeled data, we adopt Inverse Propensity Weighting (IPW)~\cite{glynn2010introduction} theory which aims to force the model to pay same attention to popular ones and tail ones in SSL models. Specifically, we use the prior knowledge of marginal distribution to adjust the optimization objective from P($\mathcal{Y}$|$\mathcal{X}$) to P($\mathcal{X}$|$\mathcal{Y}$) (Maximizing the probability of each $x \in \mathcal{X}$ with different $y \in \mathcal{Y}$ ) for unbiased learning.  
Thus, the negative impact caused by label distribution shift can be mitigated.
Consequently, the two subroutines mutually boost each other with respect to their common goal for better SSDA.



Our contributions are three-fold: (1) We formalize the DA problem using causality and propose the explicitly invariant concept learning paradigm for robust DA. (2) To unleash the power of cross-domain data, we develop a collaboratively debiasing learning framework that effectively reduces the domain gap to enforce invariant prediction. (3) We extensively evaluate the proposed \method{}. The empirical results show that it outperforms SOTA approaches on the commonly used benchmarks.


\section{Domain Adaptation through Causal Lenses: Finding the Devil}
\label{sec:da}
\begin{wrapfigure}[18]{r}{0.3\textwidth}
    \begin{center}
    \vspace{-0.7cm}
        \includegraphics[width=0.28\textwidth]{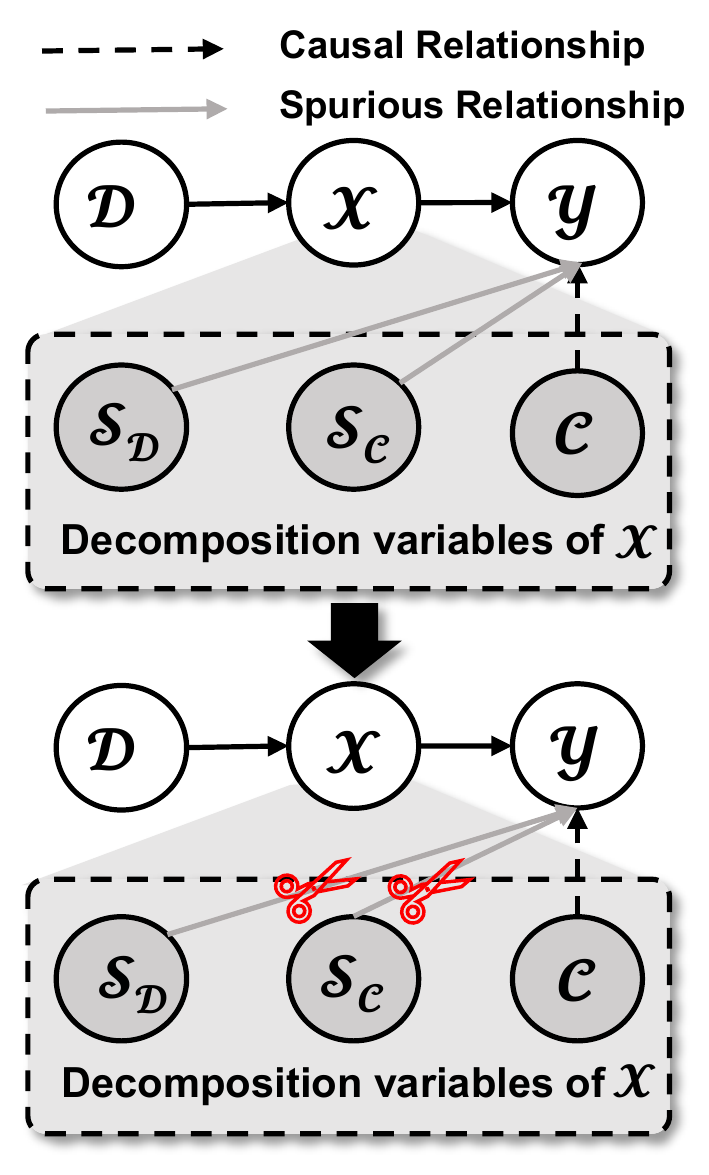}
        \caption{Causal graph of DA.}
        \label{fig:cg_1}
    \end{center}
\end{wrapfigure}
We shall start by grounding the domain adaptation (DA) in a causal framework to illustrate the key challenges of cross-domain generalization. As discussed in introduction, given data $\mathcal{X}$ and their labels $\mathcal{Y}$, the main difficulty of DA is that the extracted representation from $\mathcal{X}$ is no longer a strong visual cue for sample label in another domain. To study this issue in-depth, we first make the following assumption:

\noindent\textbf{Assumption 1 (Disentangled Variables). } \emph{ Data $\mathcal{X}$ can be disentangled into concept $\mathcal{C}$, cross-domain style $\mathcal{S}_\mathcal{C}$ and intra-domain style $S_I$ variables which are mutually independent,  i.e., $\mathcal{X}=(\mathcal{C},\mathcal{S}_\mathcal{C},\mathcal{S}_\mathcal{I})$, where $ \quad \mathcal{C} \perp \!\!\! \perp \mathcal{S}_\mathcal{C} \perp \!\!\! \perp \mathcal{S}_\mathcal{I}$. Only concept $\mathcal{C}$ is relevant for the true label $\mathcal{Y}$ of $\mathcal{X}$, i.e., style changing is concept-preserving. }

Under this assumption,  we abstract the DA problem into a causal graph (Figure~\ref{fig:cg_1} ).
In this figure, $\mathcal{D}$ represents the Domain (\emph{e.g.}, $\mathcal{D}_\mathcal{S}$ or $\mathcal{D}_\mathcal{T}$), while $\mathcal{S}_\mathcal{I}$ (\emph{e.g.}, different appearance of concept in same domain) and $\mathcal{S}_\mathcal{C}$ (\emph{e.g.}, different background of concept cross-domain) are the nuisance variables that confound $\mathcal{Y}$.
The absence of any style changing is irrelevant for true label $\mathcal{Y}$. $\mathcal{C}$ is the invariant concept which contains directly causal relationships with $Y$. Therefore, the causal graph reveals the fundamental reasons for distinguishing issues across domains, \emph{i.e.}, the cross/intra-domain style serves as the confounding variables that influence the $\mathcal{X} \rightarrow \mathcal{Y}$.
\begin{equation}
\begin{aligned}
& P(\mathcal{Y}|\mathcal{C}, \mathcal{D} = \mathcal{D}_\mathcal{S}) = P(\mathcal{Y}|\mathcal{C},\mathcal{D} = \mathcal{D}_\mathcal{T}) \quad and \quad  
P(\mathcal{Y}|\mathcal{S},\mathcal{D} = \mathcal{D}_\mathcal{S}) \neq  P(\mathcal{Y}|\mathcal{S}, \mathcal{D} = \mathcal{D}_\mathcal{T})  \\&
 \Longrightarrow P(\mathcal{Y}|\mathcal{X}, \mathcal{D} = \mathcal{D}_\mathcal{S}) \neq P(\mathcal{Y}|\mathcal{X}, \mathcal{D} = \mathcal{D}_\mathcal{T}), \quad \forall \mathcal{S} \in \{ \mathcal{S}_\mathcal{C}, \mathcal{S}_\mathcal{I} \}, 
 \label{equ_2}
\end{aligned}
\end{equation}
\emph{The ``devil'' for DA problem could be style confounders $\mathcal{S}_\mathcal{C}$ and $\mathcal{S}_\mathcal{I}$}
in that they prevent the model from learning the concept-invariant causality $\mathcal{X} \rightarrow \mathcal{Y}$\footnote{While this assumption may not be true in all settings, we believe that the single image classification can be approximated by this assumption. More discussion about this assumption is in the appendix.}. From the causal theoretical view, such confounding effect can be eliminated by statistical learning with causal intervention~\cite{pearl2000models}. 
Putting all these observations together, we now state the main theorem of the paper.


\textbf{Theorem 1 (Causal Intervention)}. \emph{ Under the causal graph in Figure~\ref{fig:cg_1} and Assumption 1, we can conclude that under this causal model, performing interventions on $\mathcal{S}_\mathcal{C}$ and $\mathcal{S}_\mathcal{I}$ does not change the $P(\mathcal{Y}|\mathcal{X})$. Thus, in DA problem, the causal effect $P(\mathcal{Y}|do(\mathcal{X})\footnote{$P(\mathcal{Y}| do(\mathcal{X}), \mathcal{D}=\mathcal{D}_\mathcal{T}$) uses the \emph{do}-operator~\cite{glymour2016causal}. Given random variables $\mathcal{X}$, $\mathcal{Y}$, we write $P(\mathcal{Y}=y| {do} (\mathcal{X}=x))$ to indicate the probability of $Y=y$  when we intervene and set $\mathcal{X}$ to be $x$.  }, \mathcal{D} = \mathcal{D}_\mathcal{T})$ can be computed as:}
\begin{equation}
\begin{aligned}
&P(\mathcal{Y}|do(\mathcal{X}), \mathcal{D} = \mathcal{D}_\mathcal{T})=P(\mathcal{Y}|do\!\!\!\!\!\!\!\!\underbrace{(\mathcal{C}, \mathcal{S}_\mathcal{C}, \mathcal{S}_\mathcal{I})}_{\rm{Disentangled\ Variables}}\!\!\!\!\!\!\!\!,\mathcal{D} = \mathcal{D}_\mathcal{T}) =\sum_{\mathcal{D} \in \{\mathcal{D}_\mathcal{S}, \mathcal{D}_\mathcal{T}\}} \sum_{\hat{s}_\mathcal{C}\sim \mathcal{S}_\mathcal{C}}  \sum_{\hat{s}_\mathcal{I} \sim \mathcal{S}_\mathcal{I}}  \\& P(\mathcal{Y}|\mathcal{C}, \hat{s}_\mathcal{C},\hat{s}_\mathcal{I}, \mathcal{D}) P(\mathcal{C},\hat{s}_\mathcal{C},\hat{s}_\mathcal{I}, \mathcal{D}) \approx  \sum_{\hat{x} \sim \hat{\mathcal{X}}}P(\mathcal{Y}|\mathcal{X},\hat{\mathcal{X}}={\hat{x}}) P(\mathcal{X},\hat{\mathcal{X}}={\hat{x}}), 
 \label{equ_3}
\end{aligned}
\vspace{-0.2cm}
\end{equation}
where $\hat{\mathcal{X}}$ are the invariant causal factors with the same concepts of $\mathcal{X}$ but contain different cross/intra-domain styles, \emph{i.e.}, invariant concept-aware samples. Realistically, $\hat{\mathcal{X}}$ is often a large set due to the multiple style combinations. This may block the model's computational efficiency according to Eq.~\ref{equ_3} and hard to obtain such numerous causal factors. However, it is non-trivial to personally determine the $\hat{\mathcal{X}}$ size to study the deconfounded effect. We employ a compromise solution that significantly reduces the $\hat{\mathcal{X}}$ size to a small number for causal intervention. 

\section{\method{}:  Causal and Collaborative Proxy-tasks Learning}
This section describes the \method{} for Semi-Supervised Domain Adaptation (SSDA) based on the studied causal and collaborative learning. We shall present each module and its training strategy.

\subsection{Problem Formulation}
In the problem of SSDA, we have access to a set of labeled samples $\mathcal{S}_l= \{(x^{(i)}_{sl}, y^{(i)}_{sl})\}_{i=1}^{\mathcal{N}_s}$  i.i.d from source domain $\mathcal{D}_\mathcal{S}$. And the goal of SSDA is to adapt a learner to a target domain $\mathcal{D}_\mathcal{T}$, of which the training set consists of two sets of data: a set of unlabeled data  $\mathcal{T}_u= \{(x^{(i)}_{tu})\}_{i=1}^{\mathcal{N}_u}$ and a small labeled set  $\mathcal{T}_l= \{(x^{(i)}_{tl}, y^{(i)}_{tl})\}_{i=1}^{\mathcal{N}_l}$.
Typically, we have $\mathcal{N}_l \leq \mathcal{N}_u$ and $\mathcal{N}_l \ll \mathcal{N}_s$. We solve the problem by decomposing the SSDA task into two proxy subroutines: Invariant Concept Learning  (ICL) and Collaboratively Debiasing Learning (CDL). 
Such subroutines are designed to seek a robust learner $\mathcal{M}(\cdot;\Theta)$ which performs well on test data from the target domain:
\begin{equation}
\begin{aligned}
\!\!\!\!\!\!\underbrace{\mathcal{M}(\cdot;(\Theta_{\mathcal{I}},\Theta_{\mathcal{C}}))}_{\rm{Learner\ \method{}}}:  \underbrace{\mathcal{M}_{\mathcal{I}}((\mathcal{\hat{S}}_{l},\mathcal{\hat{T}}_{l}, \mathcal{\hat{T}}_{u})|(\mathcal{S}_l, \mathcal{T}_u, \mathcal{T}_l));\Theta_{\mathcal{I}})}_{\rm{ICL\ Proxy\ Subroutine}}  \leftrightarrow
\underbrace{\mathcal{M}_{\mathcal{C}}((\mathcal{T}_p|(\mathcal{S}_{l},\mathcal{\hat{S}}_{l}, \mathcal{T}_l, \mathcal{\hat{T}}_l, \mathcal{T}_u,\mathcal{\hat{T}}_u ));\Theta_{\mathcal{C}})}_{\rm{CDL\ Proxy\ Subroutine}}
 \label{equ_1}
\end{aligned}
\end{equation}
where $\mathcal{M}_{I}$ and $\mathcal{M}_{C}$ indicate the ICL model parameterized by $\Theta_{\mathcal{I}}$ and the CDL model parameterized by $\Theta_{\mathcal{C}}$ respectively. 
In ICL proxy,  $\mathcal{M}_{I}(\cdot;\Theta_{I})$ learns the causal factors ($\mathcal{\hat{S}}_{l},\mathcal{\hat{T}}_{l}, \mathcal{\hat{T}}_{u}$) for $\mathcal{D}_\mathcal{S}$ and $\mathcal{D}_\mathcal{T}$ in unsupervised learning paradigm, aiming to generate the invariant causal factors and  use Eq.~\ref{equ_3} to remove the confounding effect. 
In CDL aspect, we construct two pseudo labeling-based SSL techniques: ($\mathcal{S}_l,\mathcal{\hat{S}}_l)  \rightarrow \mathcal{T}_u$ and ($\mathcal{T}_l,\mathcal{\hat{T}}_l) \rightarrow \mathcal{T}_u$, aiming at utilizing all the training data possible to bridge the feature discrepancy under the premise of invariant concept learning.




\subsection{Invariant Concept Learning Proxy}
As we discussed in Section~\ref{sec:da}, the key to robust DA is to eliminate the spurious correlations between  styles ($\mathcal{S}_\mathcal{C}$ and $\mathcal{S}_\mathcal{I}$) and label $\mathcal{Y}$. To tackle this problem,  we propose an \emph{approximate} solution to kindly remove the confounding effect induced by  $\mathcal{S}_\mathcal{C}$ and $\mathcal{S}_\mathcal{I}$. In detail, we develop the two invariant causal factor generators that can produce the causal factors $\hat{X}$ with $\mathcal{C}$. Next, we propose the Invariant Concept Learning (ICL) loss function, which forces the backbone (\emph{e.g.}, ResNet-34 ~\cite{he2016deep} ) to focus on learning concepts that are invariant across a set of domains. 
\subsubsection{Invariant Causal Factor Generator}
Achieving the invariant concept-aware $\hat{X}$ is challenging due to the fact that supervised signals are missing or expensive to obtain. 
Thus, we resort to the unsupervised learning paradigm, designing two causal factor generators $C^{fg}(\cdot)$=$C_\mathcal{C}^{fg}(\cdot)$ (cross-domain) and $C_\mathcal{I}^{fg}(\cdot)$ (intra-domain) to
 achieve $\hat{\mathcal{X}}$ for $\mathcal{D}_\mathcal{S}$ and $\mathcal{D}_\mathcal{T}$ without the reliance on the supervised signals. Take $\mathcal{D}=\mathcal{D}_\mathcal{S}$  as an example, the invariant causal factors of 
$\mathcal{S}_l$ is given by $\mathcal{\hat{S}}_l=\{\mathcal{\hat{S}}_l^t, \mathcal{\hat{S}}_l^s\}=\{C_\mathcal{C}^{fg}(\mathcal{S}_l), C_\mathcal{I}^{fg}(\mathcal{S}_l)\}$ w.r.t  $\mathcal{S}_\mathcal{C}$  and  $\mathcal{S}_\mathcal{I}$:

\noindent\textbf{Cross-domain Causal Factor. }
 $\mathcal{\hat{S}}_l^t$ are generated by $\mathcal{N}_g$ GAN-based techniques~\cite{creswell2018generative}, enabling the source concept to be preserved during the cross-domain conversion process. By considering the huge domain discrepancy, we optimize the style transfer loss as follows:
 \begin{equation}
 \begin{aligned}
&\mathop{\rm min}_{G^{k}_{st}} \mathop{\rm max}_{D_t^{k}} {\mathcal{L}_{st}^k}(\cdot;\Theta_{F}) = {\mathbbm{E}}_{x_{sl} \sim \mathcal{S}_l, x_{t}\sim [\mathcal{T}_u;\mathcal{T}_l]}  [{\rm log} D_t^{k} (x_{t})+{\rm log}  (1-D_t^k(G^{k}_{st}(x_{sl}))) \\&
+ \mathcal{L}_{cyc}^k(x_{sl},x_{t};\Theta_{F})+\mathcal{L}_{idt}^k(x_{sl},x_{t};\Theta_{F}) ], k=\mathop{{\rm argmin}}_{\{i \in {1, \cdots, N_g}\}}  \mathcal{L}_{st}^i(\cdot;\Theta_{F}),
\label{eq:ccf}
\end{aligned}
\end{equation}
where $[\cdot;\cdot]$ represents the union of two inputs, $D_t$ is the discriminator to distinguish the original source of the latent vector if from $\mathcal{D}_\mathcal{T}$. $\mathcal{L}_{cyc}^k$ and $\mathcal{L}_{idt}^k$ are the cycle and identity loss~\cite{zhu2017unpaired}. $G^{k}_{st}$ is $k^{th}$ $C_\mathcal{C}^{fg}$. 
Through min-max adversarial training, the domain style-changing samples are obtained. 

\noindent\textbf{Intra-domain Causal Factor. } We utilize the image augmentations as intra-domain style interventions, \emph{e.g.}, modifying color temperature, brightness, and sharpness. We randomly adjust these image properties as our mapping function to change the intra-domain style for $\mathcal{D}_\mathcal{S}$ with invariant concept.

Thus, the invariant causal factors $\mathcal{\hat{S}}_l=\{\mathcal{\hat{S}}_l^t, \mathcal{\hat{S}}_l^s\}$ are produced. Correspondingly, for the target domain, $\mathcal{\hat{T}}_l$ and $\mathcal{\hat{T}}_u$ are also obtained in the generating learning strategy. 

\subsubsection{ICL Optimization Objective}
\label{sec:icl}
After obtaining a set of  invariant concept-aware samples $\mathcal{S}_l$ for source domain $\mathcal{D}_\mathcal{S}$, the goal of the proposed ICL can thus be formulated as the following optimization problem:
 \begin{equation}
 \begin{aligned}
& \!\!\!\!\!\!\mathop{\rm min}_{\Theta_{\mathcal{I}}^b,\Theta_{\mathcal{I}}^c} \mathcal{L}_{icl}(\cdot;(\Theta_{\mathcal{I}}^b,\Theta_{\mathcal{I}}^c))= {\mathbbm{E}}_{(\tilde{x}_{sl},y_{sl})  \sim [\mathcal{S}_l,\mathcal{\hat{S}}_l^t,\mathcal{\hat{S}}_l^s]}[\mathcal{L}_{cls}(\Phi({\tilde{x}_{sl}} ;\Theta_{\mathcal{I}}^b),{y}_{sl};\Theta_{\mathcal{I}}^c)+\lambda_{ir}\cdot   
 \mathcal{L}_{ir}(\cdot;\Theta_{\mathcal{I}}^b)]\\&
\!\!\!s.t. \quad \Theta_{\mathcal{I}}^b=\mathop{\rm{arg}\, {min}}_{\hat{\Theta_{\mathcal{I}}^b}} \sum_{{x_{sl} \sim \mathcal{S}_l}} (\sum_{\mathcal{G} \in \{\mathcal{C},\mathcal{I}\}}
   d(\Phi({x}_{sl}),f(C^{fg}_{\mathcal{G}}({x}_{sl})))+d(\Phi(C^{fg}_{\mathcal{C}}({x}_{sl})),\Phi(C^{fg}_{\mathcal{I}}({x}_{sl})))
    \label{eq:icl}
    \vspace{-0.2cm}
\end{aligned}
\end{equation}
where $\Theta_{\mathcal{I}}^b$ and $\Theta_{\mathcal{I}}^c$ are learnable parameters for the backbone and classifier, respectively. $\Phi({\tilde{x}_{sl}}  ;\Theta_{\mathcal{I}}^b)$ is the backbone extracting feature from ${\tilde{x}_{sl}}$.
$\lambda_{ir}$ is the trade-off parameter and $d(\cdot)$ is the euclidean distance between two inputs.
$\mathcal{L}_{cls}(\Phi({\tilde{x}_{sl}};\Theta_{\mathcal{I}}^b),{y}_{sl};\Theta_{\mathcal{I}}^c)$ is the cross-entropy loss for classification. To further access the concept-invariant learning effect, we develop the invariant regularization loss $\mathcal{L}_{ir}(\cdot;\Theta_{\mathcal{I}}^b)$ through a regularizer. We feed the $\mathcal{S}_l, \hat{\mathcal{S}}_l^s, \hat{\mathcal{S}}_l^t$ into the backbone network and  explicitly enforcing them have invariant prediction, \emph{i.e.}, ${\rm KL} (P(\mathcal{Y}|\mathcal{S}_l),P(\mathcal{Y}|\hat{\mathcal{S}}_l^s), P(\mathcal{\mathcal{Y}}|\hat{\mathcal{S}}_l^t)) \leq \epsilon$\footnote{Note that any distance measure on distributions can be used in place of the Kullback-Leibler (${\rm KL}$) divergence~\cite{van2014renyi}}. Such regularization is converted to an entropy minimization process~\cite{mclachlan1975iterative}, which encourages the classifier to focus on the \emph{domain-invariant concept} and downplay the \emph{domain-variant style}. The key idea of ICL similarly corresponds to the principle of \textbf{invariant risk minimization} (IRM) which aims to model the data representation for invariant predictor learning. More discussion about IRM and ICL is in the appendix.

\subsection{Collaboratively Debiasing Learning Proxy}
After invariant concept-aware samples generation, we obtain the $\mathcal{\hat{S}}_l$, $\mathcal{\hat{T}}_l$ and $\mathcal{\hat{T}}_u$. 
Next, we will elaborate on how to utilize the advantages of the extra supervised signals of target domain data $\mathcal{T}_l$ over the UDA setting. 
We introduce the Collaboratively Debiasing Learning framework (CDL)
based on the robust DA setting with causal intervention. Specifically, we construct two SSL models: $\mathcal{M}_{\mathcal{C}}^{s}(\cdot;\Theta_\mathcal{C}^s)$ \emph{w.r.t} \{$\mathcal{S}_l$, $\hat{\mathcal{S}}_l$ and $\mathcal{T}_u$\} and $\mathcal{M}_{\mathcal{C}}^{t}(\cdot;\Theta_\mathcal{C}^t)$ \emph{w.r.t} \{$\mathcal{T}_l$, $\hat{\mathcal{T}}_l$ and $\mathcal{T}_u$\} as two complementary models with the same network architecture, which can cooperatively and mutually produce the pseudo-labels for each other to optimize the parameters~\cite{chen2011co,qiao2018deep}. For instance, pseudo-label $\bar{y}$ of $x_{tu} \sim \mathcal{T}_u$ from $\mathcal{M}_{\mathcal{C}}^{s}(\cdot;\Theta_\mathcal{C}^s)$ is given by: 
 \begin{equation}
 \begin{aligned}
 \bar{y}= \mathop{\rm{arg}\, {max}}_{\tilde{y}} (P(\tilde{y}|x_{tu};\Theta_\mathcal{C}^s)>\tau_s), \  where \ \   
 \tilde{y}= \mathcal{M}_{\mathcal{C}}^{t}(x_{tu};\Theta_\mathcal{C}^t) \ \ if \ \  P(\tilde{y}|x_{tu};\Theta_\mathcal{C}^t)>\tau_t
 \label{eq:pl}
\end{aligned}
\end{equation}
where $\tau_s$ and $\tau_t$ are the predefined threshold for pseudo-label selection. We will further elaborate on the two components in CDL, namely, a debiasing mechanism and a self-penalization technique. Without loss of generality, we describe the components using one of the SSL models $\mathcal{M}_{\mathcal{C}}^{s}(\cdot;\Theta_\mathcal{C}^s)$.

\subsubsection{Confirmation Bias Eliminating Mechanism}
The ultimate objective of most SSL frameworks is to minimize a risk, defined as the expectation of a particular loss function over a labeled data distribution $(\mathcal{X},\mathcal{Y}) \sim  \mathcal{S}_l$~\cite{van2020survey}. Therefore, the optimization problem generally becomes finding ${\Theta}_\mathcal{S}^s$ that minimizes the SSL risk.
 \begin{equation}
 \begin{aligned}
 &\mathop{\rm min}_{{\Theta}_\mathcal{C}^s} \mathcal{R}(\cdot;{\Theta}_\mathcal{C}^s)= {\mathbbm{E}}_{(x_{sl},y_{sl}) \sim \mathcal{S}_l} [ Ent_s ((x_{sl},y_{sl});{\Theta_\mathcal{C}^s})] + {\mathbbm{E}}_{x_{tu}\sim  \mathcal{T}_u} [\lambda_u \cdot  Ent_u(x_{tu};{\Theta_\mathcal{C}^s})], \\&
 \quad s.t. \quad \Theta_\mathcal{C}^s=\mathop{\rm{arg}\, {max}}_{\hat{\Theta}_\mathcal{C}^s} \sum_{(x_{sl},y_{sl}) \sim \mathcal{S}_l} {\rm log} P_s(y_{sl}|x_{sl};\hat{\Theta}_\mathcal{C}^s) ]
 \vspace{-0.4cm}
\label{ssl_loss}
\end{aligned}
\end{equation}
where $\lambda_u$ is the fixed scalar hyperparameter denoting the relative weight of the unlabeled loss. $Ent_s(\cdot)$ and $Ent_t(\cdot)$ are the cross-entropy loss function for labeled data $\mathcal{S}_l$ and unlabeled data $\mathcal{T}_u$. 


\noindent\textbf{Proposition 1 (Origin of Confirmation Bias).} \emph{SSL methods estimate the model parameters $\Theta_\mathcal{C}^s$ via maximum likelihood estimation according to labeled data $(\mathcal{X},\mathcal{Y}) \sim  \mathcal{S}_l$. Thus, the confirmation bias $\mathcal{B}_c$ in SSL methods is generated from the fully observed instances, namely labeled data.}

Under this proposition, the unbiased SSL learner should be impartial for less popular data (\emph{e.g.}, tail samples $\mathcal{X}_t$) and popular ones (\emph{e.g.}, head samples $\mathcal{X}_h$), \emph{i.e.}, $P(\mathcal{X}_h|\mathcal{Y})=P(\mathcal{X}_t|\mathcal{Y})$. Inspired by the inverse propensity weighting~\cite{glynn2010introduction} theory, we get the unbiased theorem for SSL.

\noindent\textbf{Theorem 2 (Unbiased SSL Label Propagator).} The optimization parameter  for SSL model should be taken same attention for all the labeled data, 
\emph{i.e.}, turn maximizing $\sum_{x_{sl} \in \mathcal{S}_l} {\rm log} P(x_{sl}|y_{sl});\Theta_\mathcal{C}^s) $ (Complete proof in Appendix.).
 \begin{equation}
 \begin{aligned}
&\!\!{\Theta}_\mathcal{C}^s=\mathop{\rm{arg}\, {max}}_{\hat{\Theta}_\mathcal{C}^s} \!\!\!\!\!\!\sum_{(x_{sl},y_{sl})  \sim \mathcal{S}_l}\!\!\!\!\!\! {\rm log} P(y_{sl}|x_{sl};\hat{\Theta}_\mathcal{C}^s)  =\mathop{\rm{arg}\, {max}}_{\hat{\Theta}_\mathcal{C}^s}\sum_{x_{sl} \sim \mathcal{S}_l} {\rm log} P(x_{sl}|y_{sl};\hat{\Theta}_\mathcal{C}^s)\cdot 
S_{IPW}(x_{sl}, y_{sl})\\&
S_{IPW}(x_{sl}, y_{sl})=\!\!\!\!\sum_{(x_{sl},y_{sl}) \sim \mathcal{S}_l}\!\!\!\! {P(y_{sl}|x_{sl};\hat{{\Theta}}_\mathcal{C}^s)}/ ({\rm log}{P(y_{sl}|x_{sl};\Theta_C^s)-{\rm log} P(y_{sl};\hat{\Theta}_\mathcal{C}^s)}) 
\label{eq:unbias}
\end{aligned}
\end{equation}
where $S_{IPW}(\cdot)$ is the Inverse Probability Weighting score. This formula can be understood as using the prior knowledge of marginal distribution ${P(\mathcal{Y};\Theta_\mathcal{C}^s)}$ to adjust the optimization objectives for unbiased learning. To make practical use of this Eq.~\ref{equ:unbias}, we estimate ${P(\mathcal{Y};\Theta_\mathcal{C}^s,B_s,t)}$ in each mini-batch training for error backpropagation at iteration $t$ with batch size $B_s$. It is noteworthy that we use a distribution moving strategy over all the iterations to reduce the high-variance estimation between time adjacent epochs. With the gradual removal of bias from the training process, the performance gap between classes also shrinks, and both popular and rare classes can be fairly treated.

\subsubsection{Self-penalization of Individual Classifier}

We also design a self-penalization that encourages the SSL model to produce more convincing pseudo-labels for exchanging peer classifier knowledge. 
Here, the negative pseudo-label indicates the most confident label (top-1 label) predicted by the network with a confidence lower than the threshold $\tau_s$. Since the negative pseudo-label is unlikely to be a correct label, we need to increase the probability values of all other classes except for this negative pseudo-label. Therefore, we optimize the output probability corresponding to the negative pseudo-label to be close to zero. The objective of self-penalization is defined as follows:
 \begin{equation}
 \begin{aligned}
\!\!\!\!\mathop{\rm min}_{\Theta_{\mathcal{C}}^s}\mathcal{L}_{sp}(\cdot;\Theta_{\mathcal{C}}^s) = {\mathbbm{E}}_{(x_{tu},y_{tu}) \sim \hat{\mathcal{T}}_u} [\mathbbm{1} ({\rm max}(P(y_{tu}|x_{tu};{\Theta_{\mathcal{C}}^s}) < \tau_s)) \cdot y_{tu} {\rm log}(1-P(y_{tu}|x_{tu};{\Theta_{\mathcal{C}}^s}))]\!\!
\end{aligned}
\end{equation}
Such self-penalization is able to encourage the model to generate more faithful pseudo-labels with a high-confidence score, and hence improve the data utilization for better invariant learning.


\section{Experiments}
\label{sec:Experiments}

\subsection{Dataset and Setting}
\label{sec:setting}
\noindent\textbf{Benchmark Datasets.} \texttt{DomainNet} is originally a multi-source domain adaptation benchmark. 
Following~\cite{saito2019semi} in its use for SSDA evaluation, we only select 4 domains, which are Real, Clipart, Painting, and Sketch (abbr. \textbf{R}, \textbf{C}, \textbf{P} and \textbf{S}), each of which contains images of 126 categories. \texttt{Office-Home}~\cite{venkateswara2017deep}  benchmark contains 65 classes, with 12 adaptation scenarios constructed from 4 domains (\emph{i.e.}, \textbf{R}: Real world, \textbf{C}: Clipart, \textbf{A}: Art, \textbf{P}: Product). \texttt{Office}~\cite{saenko2010adapting} is a relatively small dataset contains three domains including DSLR, Webcam and Amazon (abbr. D, W and A) with 31
classes.


\noindent\textbf{Implementation Details.}  We employ the ResNet-34~\cite{he2016deep} and VGG-16~\cite{simonyan2014very} as the backbone model on \texttt{DomainNet} and \texttt{Office-Home}, respectively. We train \method{} with a SGD~\cite{bottou2010large} optimizer in all experiments.
Besides, we use an identical set of hyperparameters ($B$=24, $M_o$=0.9,  $L_r$, $\tau$=0.5, $T_{max}$=20,000,  $\lambda_s$=1, $\lambda_u$=1, $\lambda_{ir}$=0.1, $\lambda_{sp}$=0.1). The  causal factor generator  $C^{fg}_{\mathcal{C}}(\cdot)$={CycleGan~\cite{zhu2017unpaired} and $C^{fg}_{\mathcal{I}}(\cdot)$=Image Augmentation}, $\mathcal{M}_{\mathcal{C}}(\cdot;\Theta_\mathcal{C})$=Mixmatch~\cite{berthelot2019mixmatch}) \footnote{$B$, $M_o$,$L_r$ and $T_{max}$ refer to batch size,  momentum, learning rate and max iteration in SGD optimizer. The $\mathcal{M}_{\mathcal{I}}$ and $\mathcal{M}_{\mathcal{C}}$ are orthogonal to other advanced style changing and SSL methods to boost SSDA further.} across all datasets. 

\noindent\textbf{Comparison of Methods.} For quantifying the efficacy of the proposed framework, we compare CAKE with previous SOTA SSDA approaches, including \textbf{MME}~\cite{saito2019semi}, \textbf{DANN}~\cite{ganin2016domain}, \textbf{BiAT}~\cite{jiang2020bidirectional}, \textbf{APE}~\cite{kim2020attract}, \textbf{DECOTA}~\cite{yang2021deep}, \textbf{CDAC}~\cite{li2021cross} and \textbf{SSSD}~\cite{yoon2022semi}.  More details of baselines are in the appendix.

\begin{table*}
\begin{center}
\captionsetup{font={small,stretch=1.25}, labelfont={bf}}
\caption{\textbf{Accuracy(\%) comparison on \texttt{DomainNet} under the settings of 3-shot using Resnet34 as backbone networks.}  A larger score indicates better performance. Acronym of each model can be found in Section ~\ref{sec:setting}.
We color each row as the \colorbox{myred}{\textbf{best}}, \colorbox{myorange}{\textbf{second best}}, and \colorbox{myyellow}{\textbf{third best}}. $\Uparrow$ indicate the SSDA improvement of \method{} compared with the \colorbox{myorange}{\textbf{second best}} result.}
 \renewcommand{\arraystretch}{1.2}
 \resizebox{1\textwidth}{!}{
  \begin{tabular}{l||c c c c c c c|c}
   \toprule[1.5pt]
   \textbf{Method} & \textbf{R to C} & \textbf{R to P} & \textbf{P to C} & \textbf{C to S} & \textbf{S to P} & \textbf{R to S} & \textbf{P to R} & \textbf{Mean Accuracy}\\
   \hline
   \hline
   S+T &60.8 &63.6& 60.8 &55.6 &59.5& 53.3& 74.5& 61.2\\
   DANN~\cite{ganin2016domain} & 59.8 & 62.8 & 59.6 & 55.4 & 59.9 & 54.9 & 72.2 & 60.7\\
   MME~\cite{saito2019semi} & 72.2 & 69.7 & 71.7 & 61.8 & 66.8 & 61.9 & 78.5 & 68.9\\
    APE~\cite{kim2020attract} & 76.6 & 72.1 & 76.7 & 63.1 & 66.1 & 67.8 & 79.4 & 71.7\\
    SSSD~\cite{yoon2022semi} & 75.9 & 72.1 & 75.1 & 64.4 & 70.0 & 66.7 & 80.3 & 72.1\\
     DECOTA~\cite{yang2021deep} & \colorbox{myorange}{\textbf{80.4}}  &\colorbox{myyellow}{\textbf{75.2}}  &78.7 &68.6 &72.7  &\colorbox{myyellow}{\textbf{71.9}}  &81.5  & 75.6 \\
      CDAC~\cite{li2021cross}&
      
      \colorbox{myyellow}{\textbf{79.6}}   &75.1   &
      \colorbox{myyellow}{\textbf{79.3}} &\colorbox{myyellow}{\textbf{69.9}}  &\colorbox{myyellow}{\textbf{73.4}}   &\colorbox{myorange}{\textbf{72.5}}  &\colorbox{myyellow}{\textbf{81.9}}  &\colorbox{myyellow}{\textbf{76.0}} \\
      PACL~\cite{li2020semi}& 79.0 & \colorbox{myorange}{\textbf{77.3}} & \colorbox{myorange}{\textbf{79.4}} & \colorbox{myorange}{\textbf{70.6}} & \colorbox{myred}{\textbf{74.6}} & 71.6 & \colorbox{myorange}{\textbf{82.4}} & \colorbox{myorange}{\textbf{76.4}} \\
      \toprule[1pt]
      Baseline &75.4 &71.8 &74.2 &65.9 &70.3 &70.2  &78.8 &72.4\\
      \rowcolor{gray!40} \textbf{CAKE (Ours)} & \colorbox{myred}{\textbf{83.3}} $\Uparrow$\textbf{+2.9}  &\colorbox{myred}{\textbf{77.6}}$\Uparrow$\textbf{+1.4}  &\colorbox{myred}{\textbf{80.7}}$\Uparrow$\textbf{+1.3}  &\colorbox{myred}{\textbf{72.2}}$\Uparrow$\textbf{+1.6}   &\colorbox{myorange}{\textbf{74.3 }}$\Uparrow$\textbf{-0.3}   &\colorbox{myred}{\textbf{74.5 }}$\Uparrow$\textbf{+2.0}  &\colorbox{myred}{\textbf{83.2}}$\Uparrow$\textbf{+0.8}  &\colorbox{myred}{\textbf{78.0}}$\Uparrow$\textbf{+1.6} \\
   \cline{2-9}
   \toprule[1.5pt]
  \end{tabular}}
  \label{tab:results_1}
\end{center}
\end{table*}

\subsection{Experimental Results and Analyses}
\noindent\textbf{Comparison with SOTA Methods.} Table~\ref{tab:results_1}, 2 and 3 (in Appendix) summarize the quantitative three-shot results of our framework and baselines on \texttt{DomainNet},  \texttt{Office-Home} and \texttt{Office}. More results and analysis can be found in the Appendix. In general, irrespective of the adaptation scenario, \method{} achieves the best performance on almost all the metrics to SOTA on the two datasets. In particular, \method{} outperforms other baselines in terms of Mean Accuracy by a large margin (\texttt{DomainNet}: \textbf{\underline{1.6\% $\sim$ 17.3\%}},  \texttt{Office-Home}: \textbf{\underline{3.3\% $\sim$ 9.7\%}} and \texttt{Office}: \textbf{\underline{3.8\% $\sim$ 12.0\%}}) for SSDA task. 
Notably, our baseline, a simplified variant of \method{} without causal intervention and debiasing operation also obtained comparable results compared with SOTA (-3.6\%). These results both benefit from the carefully designed ICL and CDL  proxy subroutines that demonstrate the superiority and generalizability of our proposed model.

\noindent\textbf{Individual Effectiveness of Each Component.} We conduct an ablation study  to illustrate the effect of each  component in Table~\ref{tab:result_3}, which indicates the following: 
Causal Inference is critical to boost SSDA (Row 5 \emph{vs.} Row 6), which significantly contributes 2.4\% and 1.9\% improvement on \texttt{DomainNet} and \texttt{Office-Home}, respectively. 
Meanwhile,  Row 1 indicates that it suffers from noticeable performance degradation without the bias-removed mechanism (Row 1) (-1.3\% and -1.8\%). 
Furthermore, the results of Row 3 and Row 4 severally show the performance improvement of the Invariant Regularization ($\mathcal{L}_{ir}$) and Self-penalization ($\mathcal{L}_{sp}$). 
Summing up, We can observe that the improvement of using either module alone is distinguishable. Combining all the superior components, our \method{} exhibits steady improvement over the baselines.

\begin{figure*}[t]
\includegraphics[width=1\textwidth]{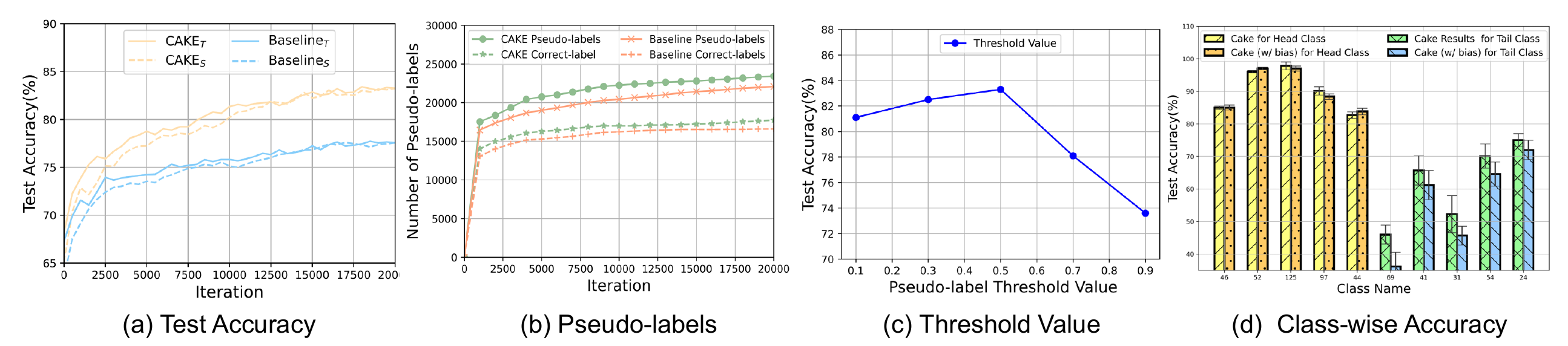}
\centering\caption{ \textbf{Analysis of cross-domain data utilization and debiasing mechanism of \method{}.} (a) and (b) depict the top-1-accuracy and correct pseudo-labels of \method{} and baseline within the first 200K iterations. (c) Cake's sensitivity to pseudo-label threshold $\tau$.  (d) demonstrates the class-wise accuracy for head and tail classes in dataset produced by \method{} (w/o)/(w/) \emph{confirmation bias}.  }
\vspace{-0.5cm}
\label{fig:analysis_1}
\end{figure*}

\noindent\textbf{Maximally Cross-domain Data Utilization.} Here, we evaluate the effectiveness of data utilization of the proposed method.   Figure~\ref{fig:analysis_1} (a) and (b) show the comparison between \method{} and baseline with respect to the top-1-accuracy, accuracy and number pseudo-labels on \texttt{DomainNet} (Real $\rightarrow$ Clipart). 

Subscript $\mathcal{S}$ and $\mathcal{T}$ represent the learned trained on source domain $\mathcal{D}_\mathcal{S}$ or target domain $\mathcal{D}_\mathcal{T}$. During the learning iteration, we observe that the accuracy of \method{} increases much faster and smoother than baseline, and outperforms baseline by a large margin of accuracy. \method{} also produces more convincing pseudo-labels than baseline. These pseudo-labels can assist the SSDA in performing global domain alignment to decrease the intra-domain discrepancy for robust invariant concept learning. Apart from learning visualization, we also investigate the \method{}'s sensitivity to the confidence threshold $\tau$ for assigning pseudo-labels. Figure~\ref{fig:analysis_1} (c) empirically provides an appropriate threshold, \emph{i.e.}, $\tau$=0.5,  either increasing or decreasing this value results in a performance decay. 
What's more, we conducted the \emph{cooperation vs. solo} ablation that verifies the power of collaborative learning in Table~\ref{tab:vs}. The detached SSL model performs worse, demonstrating that training two correlated models achieves better adaptation as opposed to only aligning one of them, because collaborative learning allows both models to learn common knowledge from different domains that in turn facilitates invariant learning. 
\begin{wraptable}[10]{r}{0.4\textwidth}
    \begin{center}
    \vspace{-0.4cm}
 \captionsetup{font={small,stretch=1.25}, labelfont={bf}}
\caption{Ablation study that showcases the impact of individual module.}
 \renewcommand{\arraystretch}{1.2}
   \resizebox{0.4\textwidth}{!}{
  \begin{tabular}{l||c c }
   \toprule[1.5pt]
   \textbf{Method} & \textbf{DomainNet} & \textbf{ Office-Home}\\
   \hline
   \hline
   -IWP Score& 75.3 &73.6\\
   -Invariant Regularization& 76.7 &74.9\\
   -Self-penalization& 77.0 &74.8\\
    -Causal Intervention& 75.2 &73.5\\
   \toprule[1pt]
   \rowcolor{gray!40} \textbf{CAKE (Ours)}& \textbf{77.6} & \textbf{75.4}\\
   \toprule[1.5pt]
  \end{tabular}}
  \label{tab:result_3}
        \label{tab:aba}
    \end{center}
\end{wraptable}
The aforementioned observation and analysis verify the effectiveness of \method{} in being able to deeply mine the potential of cross-domain data, thereby achieving the SSDA improvement.

\noindent\textbf{Effect of Confirmation Bias Eliminating.} To build insights on the unbiased SSL in \method{}, we perform an in-depth analysis of the bias-eliminating mechanism in Figure~\ref{fig:analysis_1}(d).
In this experiment, we randomly select 10 classes (5 head and 5 tail).
The results suggest that \method{} and its variant \method{} (w/ bias) obtain a comparable performance on the head class. However,  \method{} (w/ bias)  fails to maintain the consistent superiority on the tail class while our approach does.
(\emph{e.g.}, tail class 69, \method{}: 46.0\% , \method{} (w/ bias) : 36.2\%). 
This phenomenon is reasonable since \method{}  maintains unbiasedness to each class-wise sample by maximizing $P(\mathcal{X}|\mathcal{Y})$. 
\begin{wraptable}[7]{r}{0.38\textwidth}
    \begin{center}
    \vspace{-0.6cm}
 \captionsetup{font={small,stretch=1.25}, labelfont={bf}}
\caption{\textbf{Results of cooperation \emph{vs.} solo.}}
 \renewcommand{\arraystretch}{1.2}
   \resizebox{0.38\textwidth}{!}{
   \centering
  \begin{tabular}{l||c c }
   \toprule[1.5pt]
   \textbf{Method} & \textbf{Domain} & \textbf{ Office-Home}\\
   \hline
   \hline
   $\mathcal{M}_{\mathcal{C}}^{s}$& 70.6 &67.4\\
    $\mathcal{M}_{\mathcal{C}}^{t}$& 68.3 &65.8\\
       \toprule[1pt]
      \textbf{CAKE (Ours)}& \textbf{77.6} & \textbf{75.4}\\
   \toprule[1.5pt]

  \end{tabular}}
        \label{tab:vs}
    \end{center}
\end{wraptable}
As the labeled/unlabeled data share the same class distribution, the accuracy of the tail class can be improved. 
In contrast, \method{} (w/ bias) focuses more on the head class, which results in an unbalanced performance for all categories. 
These results empirically verified our theoretical analysis and the robustness of the debiasing mechanism, which provides a reliable solution that guarantees the mutual data knowledge to be exchanged from source and target aspects.

\begin{figure*}[t]
\includegraphics[width=0.9\textwidth]{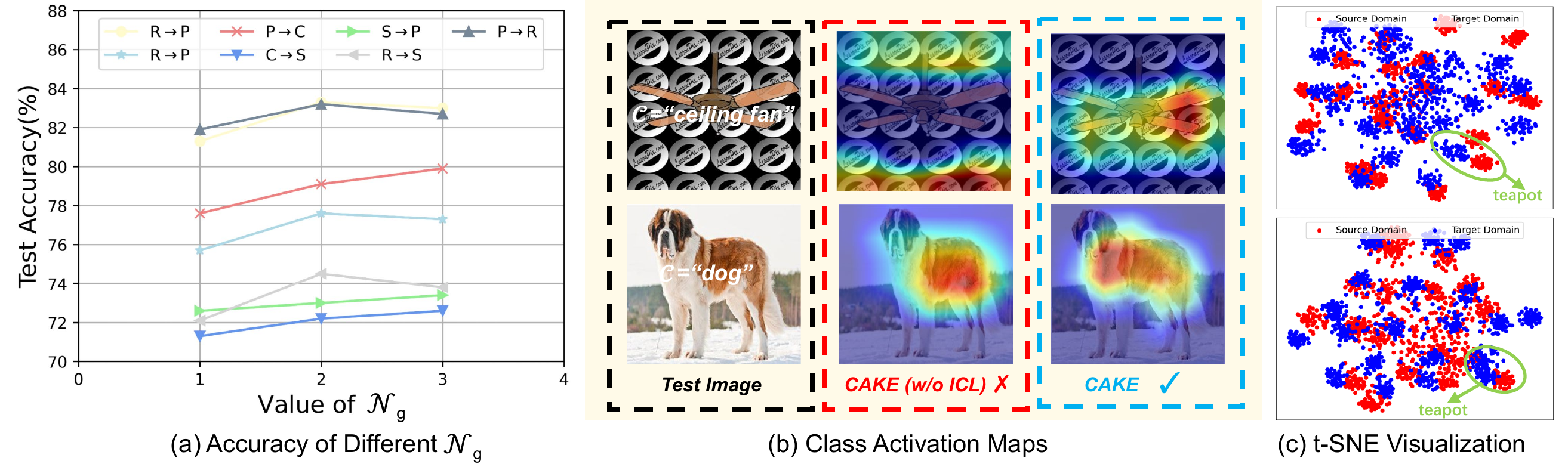}
\vspace{-0.2cm}
\centering\caption{ \textbf{In-depth analysis of \method{}.} (a) is the plot of invariant causal factor number $\mathcal{N}_g$ against accuracy(\%). (b) Grad-CAM results of CAKE and CAKE(w/o ICL). (c) t-SNE plot of features. }
\vspace{-0.6cm}
\label{fig:analysis_2}
\end{figure*}

\noindent\textbf{Number of Invariant Causal Factors.} Figure~\ref{fig:analysis_2}(a) reports the SSDA results of different numbers of Invariant Causal Factors (ICFs) $\hat{\mathcal{X}}$ ($2 \times \mathcal{N}_g$) on \texttt{DomainNet}. Across all scenes, the best performance is usually achieved with $\mathcal{N}_g$ = 2, except for P $\rightarrow$ C. This ablation proves the ICFs of $\hat{\mathcal{X}}$ can be learned from a set of limited style-changing samples. Appropriately using these ICFs to conduct the deconfounded operation can effectively improve the  SSDA performance.

\noindent\textbf{Grad-CAM Results of Causal Intervention.} We systematically present the explicit benefits of the  invariant concept learning (ICL). Figure~\ref{fig:analysis_2}(b) visualizes the most influential part in prediction generated from Grad-CAM~\cite{selvaraju2017grad}. It's rather clear to see that \method{} appropriately captures the invariant part of the concept while  \method{}(w/o CI) failed. We also analyze the reason why \method{} performs better in these cases. For instance, the concept $C$=``celling fan'' has a complicated background, \emph{i.e.}, style $\mathcal{S}$=``cluttered''.
Without causal intervention, \method{} (w/o ICL) tends to focus on the irrelevant information of style $\mathcal{S}$= ``cluttered", therefore predicting the wrong class. On the contrary, \method{} can attend to the vital image regions by learning the invariant concept $\mathcal{C}$= "celling fan" through the deconfounded mechanism.

\noindent\textbf{Cross-domain Feature Alignment.} We employ the t-SNE~\cite{van2008visualizing} to visualize the feature alignment before/after training of adaptation scenarios $\mathcal{D}_\mathcal{S}$=``Real'' and $\mathcal{D}_\mathcal{T}$=``Clipart'' on \texttt{DomainNet}. We randomly select 1000 samples (50 samples per class). Our invariant concept learning focuses on making $\mathcal{X}_\mathcal{S}$ and $\mathcal{X}_\mathcal{T}$ alike. It can be observed that as the model optimization progresses, \emph{e.g.}, $\mathcal{C}$=``teapot'', the target features gradually converge toward target cluster cores. Each cluster in the target domain also gradually moves closer to its corresponding source cluster cores, showing a cluster-wise feature alignment effect. This provides an intuitive explanation of how our \method{} alleviates the domain shift issue.
\vspace{-0.3cm}


\section{Related Work}
\vspace{-0.3cm}
\textbf{Semi-supervised Domain Adaptation.}  Semi-supervised domain adaptation (SSDA)~\cite{saito2019semi, qin2020opposite, jiang2020bidirectional, li2020online, kim2020attract,li2021cross,yoon2022semi} address the
domain adaptation problem where some target labels are available.
However, these techniques mainly rely on the two domain shift assumptions of \emph{Covariate Shift} and \emph{Conditional Shift} to conduct SSDA. Such assumptions present intuitive solutions but lack a solid theoretical explanation for the effectiveness of SSDA, which hinders their further development. Thus we develop the \method{}, which decomposes the SSDA as two proxy subroutines with causal theoretical support and reveals the fundamental reason of the two domain shift assumptions. 

\textbf{Invariant Risk Minimization.}  Recently, the notion of invariant prediction has emerged as an important operational concept in the machine learning field, called IRM~\cite{rosenfeld2020risks,arjovsky2019invariant}. IRM proposes to use group structure to delineate between different environments where the aim is to minimize the classification loss while also ensuring that the conditional variance of the prediction function within each group remains small. In DA, this idea can be studied by learning classifiers that are robust against domain shifts~\cite{li2021learning} but still has the \emph{Covariate Shift} issue. Therefore, we propose the \method{} that enforces the model to learn the local disentangled invariant-concepts rather than the global invariant-features across domains, thus facilitating the SSDA.

\textbf{Causality in DA.} There are some causality study in DA community. ~\cite{glynn2010introduction} considered domain adaptation where both the distribution of the covariate and the conditional distribution of the target given the covariate change across domains. ~\cite{gong2016domain} consider the target data causes the covariate, and an appropriate solution is to find conditional transferable components whose conditional distribution given the target is invariant after proper location-scale transformations, and estimate the target distribution of the target domain. Different from the two causal DA handle the DA task that only deals with the \emph{Conditional Shift} issue, we also consider the \emph{Covariate Shift}, which presents a improved IRM view for SSDA. 
\vspace{-0.3cm}

\section{Conclusion}

 We first propose a causal framework to pinpoint the causal effect of disentangled style variables, and theoretically explain what characteristics should a robust domain adaptation model have. We next discuss the maximal training data utilization and present a collaboratively debiasing learning framework to make use of the training data to boost SSDA effectively. We believe that \method{} serves as a complement to existing literature and provides new insights to the domain adaptation community.



\bibliography{iclr2023_conference}
\bibliographystyle{iclr2023_conference}
\newpage
\end{document}